\documentclass[letterpaper]{article} % DO NOT CHANGE THIS
\usepackage{aaai25}  % DO NOT CHANGE THIS
\usepackage{times}  % DO NOT CHANGE THIS
\usepackage{helvet}  % DO NOT CHANGE THIS
\usepackage{courier}  % DO NOT CHANGE THIS
\usepackage[hyphens]{url}  % DO NOT CHANGE THIS
\usepackage{graphicx} % DO NOT CHANGE THIS
\usepackage{amsmath}
\usepackage{amssymb}
\usepackage{natbib}  % DO NOT CHANGE THIS AND DO NOT ADD ANY OPTIONS TO IT
\usepackage{caption} % DO NOT CHANGE THIS AND DO NOT ADD ANY OPTIONS TO IT
\usepackage{subfigure}
\usepackage{algorithm}
\usepackage{algorithmic}
\usepackage{booktabs}
\usepackage{newfloat}
\usepackage{bm}
\usepackage{listings}
\usepackage{multirow}
\DeclareCaptionStyle{ruled}{labelfont=normalfont,labelsep=colon,strut=off} % DO NOT CHANGE THIS
\floatstyle{ruled}
\newfloat{listing}{tb}{lst}{}
\floatname{listing}{Listing}

\title{A Plug-and-Play Bregman ADMM Module for \\Inferring Event Branches in Temporal Point Processes}
\author{
    Qingmei Wang\textsuperscript{\rm 1}, Yuxin Wu\textsuperscript{\rm 1}, Yujie Long\textsuperscript{\rm 2}, Jing Huang\textsuperscript{\rm 2}, \\Fengyuan Ran\textsuperscript{\rm 2}, Bing Su\textsuperscript{\rm 1,3}, Hongteng Xu\textsuperscript{\rm 1,3}\thanks{Corresponding author: hongtengxu@ruc.edu.cn}
}
\affiliations{
    \textsuperscript{\rm 1}Gaoling School of Artificial Intelligence, Renmin University of China\\
     \textsuperscript{\rm 2}School of Cyber Science and Engineering, Wuhan University\\
    \textsuperscript{\rm 3}Beijing Key Laboratory of Big Data Management and Analysis Methods\\
   
}

\usepackage{bibentry}
\begin{document}

\maketitle

\begin{abstract}
An event sequence generated by a temporal point process is often associated with a hidden and structured event branching process that captures the triggering relations between its historical and current events. 
In this study, we design a new plug-and-play module based on the Bregman ADMM (BADMM) algorithm, which infers event branches associated with event sequences in the maximum likelihood estimation framework of temporal point processes (TPPs). 
Specifically, we formulate the inference of event branches as an optimization problem for the event transition matrix under sparse and low-rank constraints, which is embedded in existing TPP models or their learning paradigms.
We can implement this optimization problem based on subspace clustering and sparse group-lasso, respectively, and solve it using the Bregman ADMM algorithm, whose unrolling leads to the proposed BADMM module. 
When learning a classic TPP (e.g., Hawkes process) by the expectation-maximization algorithm, the BADMM module helps derive structured responsibility matrices in the E-step. 
Similarly, the BADMM module helps derive low-rank and sparse attention maps for the neural TPPs with self-attention layers.
The structured responsibility matrices and attention maps, which work as learned event transition matrices, indicate event branches, e.g., inferring isolated events and those key events triggering many subsequent events. 
Experiments on both synthetic and real-world data show that plugging our BADMM module into existing TPP models and learning paradigms can improve model performance and provide us with interpretable structured event branches. 
The code is available at \url{https://github.com/qingmeiwangdaily/BADMM_TPP}.
\end{abstract}

\section{Introduction}
\begin{figure}[t]
    \centering
    \subfigure[Event branches]{
    \includegraphics[height=4.1cm]{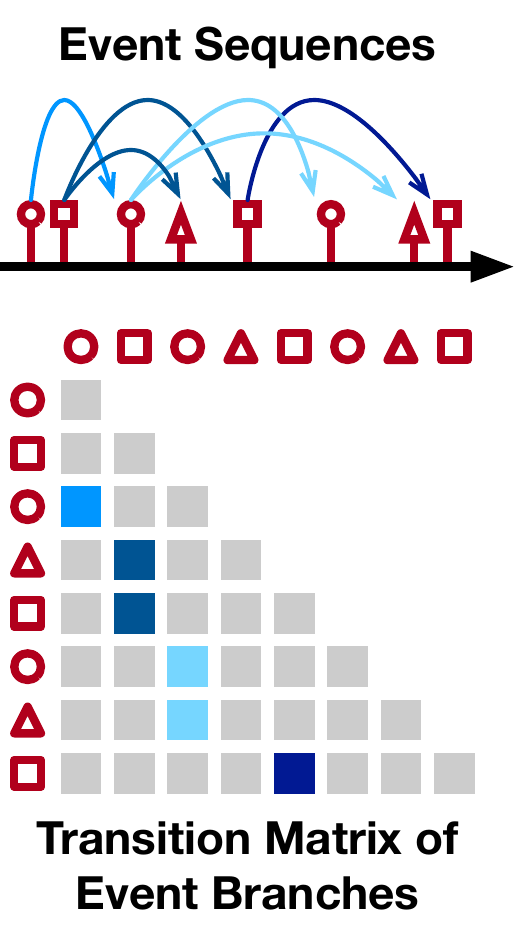}\label{fig:branch}
    }
    \subfigure[Plug-in strategies of BADMM module]{
    \includegraphics[height=4.1cm]{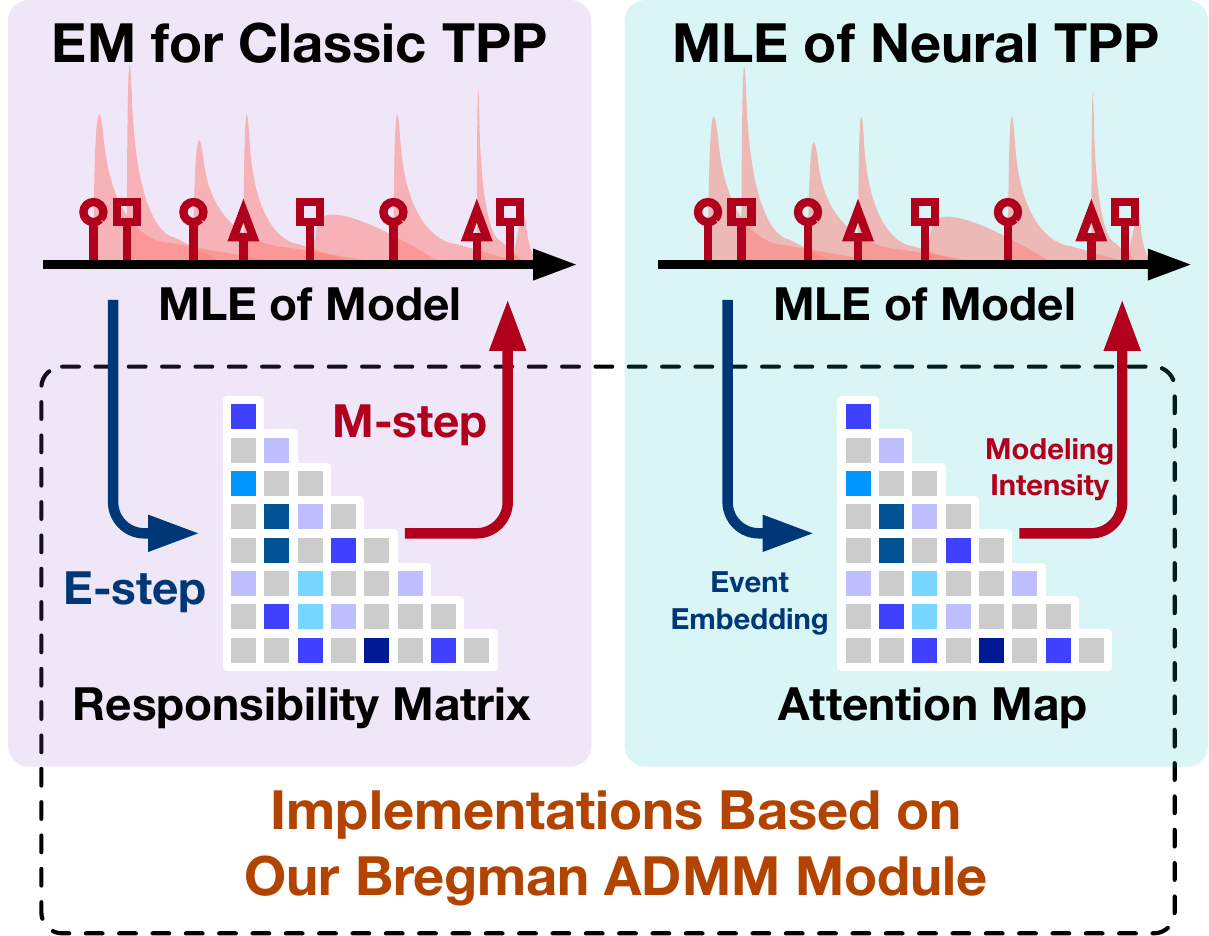}\label{fig:scheme}
    }
    \caption{(a) An illustration of event branches and the corresponding transition matrix. (b) Plug-in strategies of BADMM module for classic and neural TPPs.}
    \label{fig:1}
\end{figure}

Temporal point processes (TPPs) are powerful tools for modeling events that occur sequentially in continuous-time domain~\cite{kingman1992poisson,ross1996stochastic}. 
They have been extensively used to capture the dynamics of the events in various domains, such as earthquake prediction~\cite{lewis1979simulation}, financial analysis~\cite{bacry2015hawkes}, social network modeling~\cite{zhou2013learning2}, recommendation systems~\cite{xu2018learning}, and so on. 
Typically, for an event sequence generated by a TPP, there exists a branching process capturing the event-level triggering patterns hidden in the sequence~\cite{dion1994statistical,moller2006approximate}. 
As illustrated in Figure~\ref{fig:branch}, the branching process is often represented as a transition matrix of events, which provides insights into the causal relationships between events and helps identify the cascades of subsequent events.
Therefore, inferring such event branches from observed event sequences is crucial for us to reveal the underlying mechanism of event generation and information diffusion, e.g., identifying the source node in an information diffusion network~\cite{farajtabar2015back,zhang2018started} and controlling the diffusion of specific information~\cite{farajtabar2014shaping,farajtabar2017coevolve}.

Despite its significance, inferring event branches is challenging because they are latent variables unobservable in practice. 
For some classic TPPs like Hawkes process~\cite{zhou2013learning}, we can learn the TPPs based on observed event sequences in the expectation-maximization (EM) framework, in which the event branches are often inferred as the responsibilities of the events (i.e., the posterior probabilities of historical events given the current ones). 
For neural TPPs like transformer Hawkes process (THP)~\cite{zuo2020transformer} and self-attentive Hawkes process (SAHP)~\cite{zhang2020self}, the attention maps within the models indicate the impacts of historical events on the current ones, and thus, can be treated as evidence of event branches.
However, both methods often suffer the over-smoothness issue and lead to unstructured event branches. 
How to infer interpretable and structured event branches effectively for various TPP models is still an open problem.

In this study, we introduce a novel plug-and-play module based on the Bregman Alternating Direction Method of Multipliers (Bregman ADMM or BADMM for short) algorithm~\cite{wang2014bregman}, which helps infer event branches in the maximum likelihood estimation framework of TPPs. 
As illustrated in Figure~\ref{fig:scheme}, given the responsibility matrices of classic TPPs or the attention maps of neural TPPs, the BADMM module imposes low-rank and sparse structures on the matrices by solving a subspace clustering problem~\cite{elhamifar2013sparse} or a sparse group-lasso problem~\cite{simon2013sparse}. 
Applying the BADMM module to the EM algorithm of classic TPPs, we can optimize the responsibility matrices iteratively to guide the learning of model parameters.
For neural TPPs, we unroll the iterations of the Bregman ADMM algorithm and build the BADMM module to obtain a new attention layer. 
With the help of the BADMM module, we effectively penalize dense triggering patterns among events for both classic and neural TPPs, leading to structured event transition matrices and, accordingly, interpretable event branches. 

Through extensive experiments on both synthetic and real-world datasets, we evaluate different implementations of the BADMM module and demonstrate their effectiveness. 
Experimental results show that incorporating the BADMM module into existing TPP models improves their performance and interpretability.
In particular, the BADMM module not only enhances the predictive accuracy of TPP models but also provides valuable insights into the hidden event branching processes, e.g., identifying the isolated events within event sequences and those key events triggering multiple subsequent events. 

\section{Related Work and Preliminaries}
\subsection{Temporal Point Processes}
Temporal point process is a classic statistical tool to model the event sequences in continuous-time domain~\cite{kingman1992poisson,wang2023hierarchical}. 
Suppose that we observe an event sequence $\bm{s}=\{(t_n, c_n)\}_{n=1}^{N}$, where $(t_n, c_n)$ is the $n$-th event, $t_n\in [0,T]$ is its timestamp, and $c_n\in\mathcal{C}=\{1,...,C\}$ is its event type. 
A TPP often models the dynamics of the events by a parametric multivariate intensity function, denoted as $\bm{\lambda}(t;\theta)=\{\lambda_{c}(t;\theta)\}_{c\in\mathcal{C},t\in [0,T]}$, where
\begin{eqnarray}\label{eq:intensity}
\lambda_{c}(t;\theta)\mathrm{d}t=\mathrm{d} \mathbb{E}[N_c(t;\theta)|\mathcal{H}_t^{\mathcal{C}}],~\forall c\in\mathcal{C},
\end{eqnarray}
represents the expected instantaneous rate of the type-$c$ event happening at time $t$ given historical events.
Here, $N_c(t;\theta)$ is the counting process of the event type, which records the number of the type-$c$ events happening till time $t$, and $\mathcal{H}_t^{\mathcal{C}}=\{(t_n,c_n)\in\bm{s}|t_n<t\}$ is the history till time $t$.
The parameters of the intensity function (and the counting process) are denoted as $\theta$. 
We can learn the TPP in the maximum likelihood estimation (MLE) framework, maximizing the log-likelihood of the event sequence shown below:
\begin{eqnarray}\label{eq:loglike}
    \log L(\bm{s};\theta) = \sideset{}{}\sum_{n=1}^{N}\log\lambda_{c_n}(t_n;\theta) - \sideset{}{}\sum_{c\in\mathcal{C}}\int_{0}^{T}\lambda_{c}(s;\theta)\mathrm{d}s.
\end{eqnarray}

Classic TPP models, like Poisson process~\cite{kingman1992poisson}, Hawkes process~\cite{hawkes1971spectra}, self-correcting process~\cite{isham1979self}, and their multivariate versions~\cite{zhou2013learning,xu2016patient}, often apply manually designed intensity functions, which simplifies the learning problem but suffers a high risk of model misspecification. 
To overcome this challenge, some neural TPPs have been proposed to model intensity functions by neural networks, which can enhance model capabilities significantly. 
The early neural TPPs are built based on recurrent neural networks, e.g., RMTPP~\cite{du2016recurrent} and NHP~\cite{mei2017neural}.
Recently, some Transformer-based TPP models are proposed, e.g., SAHP~\cite{zhang2020self} and THP~\cite{zuo2020transformer}.

\subsection{Inference of Event Branches}
For the event sequence whose events have inter-dependency, there always exists a branching process associated with the corresponding TPP~\cite{moller2006approximate,farajtabar2014shaping}.
Typically, we can represent the branches of $N$ events by a lower triangular matrix, denoted as $\bm{B}=[b_{nn'}]\in [0,\infty)^{N\times N}$, where $b_{nn'}\geq 0$ if $n\geq n'$ and $b_{nn'}=0$ otherwise.  
This matrix often works as a ``transition'' matrix of events --- the nonzero $b_{nn'}$ indicates that the $n'$-th event  contributes to the happening of the $n$-th event. 
Ideally, a sparse and structured transition matrix helps reveal the underlying generative mechanisms of events~\cite{dion1994statistical,gonzalez2013bayesian,champion2022branching}, which enhances the interpretability of the TPP model.
Therefore, in this study, we aim to infer the event branch matrix $\bm{B}$ simultaneously when learning the TPP model $\theta$. 

Inferring event branches for arbitrary TPPs is challenging due to the latent nature of transition matrix. 
Focusing on Hawkes process~\cite{zhou2013learning}, whose intensity function encodes the branching process explicitly~\cite{moller2006approximate}, some attempts have been made to infer event branches.
For example, an information source identification method is proposed in~\cite{farajtabar2015back}, which traces back to information sources in diffusion networks from partially observed cascades, crucial for applications like epidemiology and social networks. 
A related study~\cite{farajtabar2017fake} designs interventions to mitigate fake news spread using point process models, demonstrating the real-world application of branch process inference in misinformation control. 
Focusing on Hawkes processes with textual covariates, the work in~\cite{zhang2018started} identifies root sources in textual conversation threads, providing insights into propagation patterns within text data. 
However, the above methods often lead to non-structured and over-dense event branches because they seldom consider imposing structural regularization on the transition matrix of events. 
Moreover, these methods only apply to Hawkes process, which cannot be extended to other complicated TPP models.

\subsection{Optimization-driven Model Design}
In this study, we aim to design a plug-and-play BADMM module that effectively infers event branches for both classic and neural TPPs.
Our work can be treated as a new attempt at optimization-driven model design. 
In this field, some methods have been made to implement implicit neural network layers based on optimization algorithms.
For example, the Sinkhormer in~\cite{sander2022sinkformers} and the ROTP in~\cite{xu2023regularized} apply Sinkhorn-scaling algorithm~\cite{cuturi2013sinkhorn} to improve attention maps and global pooling layers, respectively. 
An ADMM-based neural network is proposed to solve compressive sensing problems~\cite{yang2018admm}. 
These methods can embed optimization algorithms into model architectures or learning paradigms and, accordingly, impose structural constraints on model parameters or intermediate outputs. 
However, none of the existing methods consider inferring event branches for TPPs.

\section{Proposed Method}
\subsection{Transition Matrices within TPPs}
As aforementioned, the inference of event branches corresponds to learning structured transition matrix of events. 
This matrix is intrinsic for many existing TPP models. 

\subsubsection{Hawkes process}
The intensity function of Hawkes process encodes event branches explicitly~\cite{moller2006approximate}, which is constructed as
\begin{eqnarray}\label{eq:hp}
    \lambda_{c}(t;\theta)=\mu_c+\sideset{}{_{t_n<t}a_{cc_n}}\sum\kappa(t-t_n),
\end{eqnarray}
in which $\mu_c$ is time-invariant exogenous intensity of the type-$c$ event and $\sum_{t_n<t}a_{cc_n}\kappa(t-t_n)$ is endogenous intensity recording the accumulative impacts of historical events at time $t$. 
$\kappa(t)$, $t\geq 0$, is a predefined time-decay function.
As a result, the model parameters $\theta=\{\bm{\mu},\bm{A}\}$ include the exogenous intensity vector $\bm{\mu}=[\mu_c]\in [0,\infty)^{C}$ and the infectivity matrix $\bm{A}=[a_{cc'}]\in [0,\infty)^{C\times C}$ capturing the pairwise impacts between different event types. 

We often apply the EM algorithm~\cite{zhou2013learning} to achieve the MLE of Hawkes process. 
In the $t$-th iteration, given the current parameter $\theta^{(t)}=\{\bm{\mu}^{(t)},\bm{A}^{(t)}\}$, we update the parameter by the following two steps:
\begin{itemize}
    \item \textbf{E-step:} We construct a lower bound of the log-likelihood in~\eqref{eq:loglike} based on Jensen's inequality:
    \begin{eqnarray*}\label{eq:e_step}
    \begin{aligned}
    &\log L(\bm{s};\theta)\geq Q(\theta,\theta^{(t)}) \\
    &= \sideset{}{}\sum_{n=1}^{N} \Big( r_{nn}^{(t)} \log \frac{\mu_{c_n}}{r_{nn}^{(t)}} + \sideset{}{}\sum_{n'=1}^{n-1} r_{nn'}^{(t)} \log \frac{a_{c_n c_{n'}} \kappa(t_n - t_{n'})}{r_{nn'}^{(t)}} \Big) \\
    &\quad-\sideset{}{_{c\in\mathcal{C}}}\sum \Big( T \mu_{c} 
        + \sideset{}{_{n=1}^{N}}\sum a_{cc_n} \int_{0}^{T-t_n} \kappa(s) \mathrm{d}s \Big)
    \end{aligned}
    \end{eqnarray*}
    where $\bm{R}^{(t)}=[r_{nn'}^{(t)}]$ is the responsibility matrix in the $t$-th iteration.
    Its element is
    \begin{eqnarray}\label{eq:response}
        r_{nn'}^{(t)}=
        \begin{cases}
            \frac{\mu_{c_n}^{(t)}}{\lambda_{c_n}(t_n;\theta^{(t)})}, & n=n',\\
            \frac{a_{c_n c_{n'}}^{(t)}\kappa(t_n-t_{n'})}{\lambda_{c_n}(t_n;\theta^{(t)})}, & n>n',\\
            0, & n<n'.
        \end{cases}
    \end{eqnarray}
    which corresponds to the posterior probability of the $n'$-th event given the $n$-th event. 
    \item \textbf{M-step:} We update the model parameter by minimizing the surrogate function $Q(\theta, \theta^{(t)})$, i.e.,
    \begin{eqnarray}\label{eq:m_step}
    \theta^{(t+1)}=\arg\sideset{}{_{\theta}}\max Q(\theta,\theta^{(t)}),
    \end{eqnarray}
    which can be solved in a closed form~\cite{zhou2013learning,xu2016learning}.
\end{itemize}
We can find that the responsibility matrix in~\eqref{eq:response} works as the transition matrix of events, which is estimated by the E-step iteratively. 
Therefore, the inference of event branches corresponds to estimating a structured responsibility matrix. 

\subsubsection{Transformer-based neural TPPs}
For Transformer-based neural TPPs~\cite{zuo2020transformer,zhang2020self}, their intensity functions encode historical impacts by a self-attention mechanism.
Given the event embeddings of a sequence, denoted as $\bm{X}=[\bm{x}_1,...,\bm{x}_N]\in\mathbb{R}^{D\times N}$, the attention map is
\begin{eqnarray}\label{eq:att}
    \bm{\tilde{A}}=\sigma_{r}\Bigl(\bm{L}\odot\frac{(\bm{QX})^{\top}\bm{KX}}{\sqrt{D}}\Bigr),
\end{eqnarray}
where $\bm{Q},\bm{K}\in\mathbb{R}^{d\times D}$ maps the event embeddings to latent spaces. 
$\bm{L}=[\ell_{nn'}]$ is a lower triangular masking matrix, whose element $\ell_{nn'}=1$ if $n\geq n'$, otherwise, $\ell_{nn'}=\infty$. 
``$\odot$'' denotes the Hadamard product of matrix, and $\sigma_r$ means applying the Softmax operation to each row of input matrix.
As shown in~\cite{zuo2020transformer}, the sparsity of $\bm{\tilde{A}}$ indicates event branches --- $a_{nn'}$ with a large value implies that the $n$-th event is likely to be triggered by the $n'$-th event. 

\subsubsection{Over-smoothness issue}
The above two examples demonstrate that many TPP models estimate transition matrices intrinsically either in their learning paradigms or their model architectures. 
Unfortunately, these matrices often suffer the over-smoothness issue, i.e., having too many nonzero elements and leading to over-dense and unstructured event branches. 
For the $\bm{R}^{(t)}$ in~\eqref{eq:response}, its structure is determined by the sparsity of $\bm{\mu}$ and $\bm{A}$, which requires additional sparse regularization~\cite{xu2016learning,zhou2013learning}. 
For the $\bm{\tilde{A}}$ in~\eqref{eq:att}, its lower-triangular part is always nonzero because of using the softmax operation.
Therefore, we need to impose more structural constrains when estimating the transition matrices, which motivates the design of our BADMM module.

\subsection{BADMM Module for Structured Event Branches}
\subsubsection{Imposing sparse and low-rank structures}
Taking the $\bm{R}^{(t)}$ in~\eqref{eq:response} or the $\bm{\tilde{A}}$ in~\eqref{eq:att} as the initial transition matrix, denoted as $\bm{B}_0$, our Bregman ADMM module imposes sparse and low-rank structures on the matrix by solving the following optimization problem:
\begin{eqnarray}\label{eq:opt}
    \sideset{}{}\min_{\bm{B}\in\Omega} 
    \underbrace{\text{KL}(\bm{B}~\|~\bm{B}_0)}_{\text{Prior}}+ \lambda \underbrace{(\alpha\|\bm{B}\|_1+(1-\alpha)R(\bm{B}))}_{\text{Structural regularizer}}.
\end{eqnarray}
The feasible domain $\Omega=\{\bm{B}=[b_{nn'}] |\bm{B}\bm{1}_N=\bm{1}_N, b_{nn'}=0~\text{for}~n<n',~b_{nn'} \geq 0~\text{for}~n \geq n'\}$, ensuring $\bm{B}$ is a row-normalized lower triangular matrix.
In the objective function, the first term penalizes the KL-divergence between $\bm{B}$ and the initial $\bm{B}_0$, which requires the target transition matrix inherits the prior knowledge provided by $\bm{B}_0$ to some extent.
The second term is a joint sparse and low-rank regularization, in which the $\ell_1$-norm $\|\bm{B}\|_1=\sum_{n,n'=1}^{N}|b_{nn'}|$ enhances the sparsity of the event branch matrix and $R(\bm{B})$ penalizes the rank of $\bm{B}$. 
In this study, we consider two implementations of $R(\bm{B})$:
\begin{itemize}
    \item \textbf{Subspace clustering.} We can implement $R(\bm{B})$ as the nuclear norm, i.e., $\|\bm{B}\|_*=\sum_{n=1}^{N}\sigma_N(\bm{B})$. 
    It computes the summation of $\bm{B}$'s singular values, which encourages sparse singular values and results in a low-rank matrix. 
    As a result, the regularizer $\alpha\|\bm{B}\|_1+(1-\alpha)\|\bm{B}\|_*$ corresponds to the subspace clustering method in~\cite{elhamifar2013sparse,liu2010robust}.
    \item \textbf{Sparse group-lasso.} We can also implement $R(\bm{B})$ as the $\ell_{1,2}$ norm, i.e., $\|\bm{B}\|_{1,2}=\sum_{n=1}^{N}\|\bm{b}_n\|_2$, where $\bm{b}_n$ is the $n$-th column of $\bm{B}$. 
    This term encourages $\bm{B}$ to have a group sparse structure.
    As a result, the regularizer $\alpha\|\bm{B}\|_1+(1-\alpha)\|\bm{B}\|_*$ corresponds to the sparse group-lasso in~\cite{simon2013sparse}.
\end{itemize}
The significance of the regularization term is controlled by $\lambda >0$, and the trade-off between the sparse and low-rank terms is achieved by $\alpha\in [0, 1]$. 

\textbf{Remark.} 
It should be noted that the utilization of the regularization is motivated by the event branching structure we desire. 
As illustrated in Figure~\ref{fig:branch}, an event branching process often consists of some isolated events independent of other events and some key events that trigger subsequent events.
In addition, an event's impact always decays over time, so it can only trigger a subset rather than all subsequent events.
As a result, the corresponding event branch matrix is sparse and low-rank, and the proposed regularizer enhances such structures accordingly.

\subsubsection{Implementation details}
Applying Bregman ADMM algorithm~\cite{wang2014bregman}, we can solve the above optimization problem iteratively. 
In particular, we first rewrite~\eqref{eq:opt} in the following equivalent format:
\begin{eqnarray}\label{eq:opt2}
\begin{aligned}
    &\sideset{}{}\min_{\bm{B}, \bm{X}_1, \bm{X}_2} \text{KL}(\bm{B}\|\bm{B}_0) + \lambda(\alpha\|\bm{X}_1\|_1 + (1-\alpha)R(\bm{X}_2)) \\
    &\quad \text{s.t.}\quad\bm{B} = \bm{X}_1 = \bm{X}_2,\quad \bm{B}\in\Omega.
\end{aligned}
\end{eqnarray}
Here, $\bm{X}_1$ and $\bm{X}_2$ are two auxiliary variables helping decouple the three terms in the objective function.
Then, we can further rewrite~\eqref{eq:opt2} in its augmented Lagrangian form, i.e.,
\begin{eqnarray}\label{eq:lagrange}
\begin{aligned}
    &\sideset{}{_{\bm{Z}_1, \bm{Z}_2}}\max\sideset{}{_{\bm{B}\in\Omega, \bm{X}_1, \bm{X}_2}}\min \text{KL}(\bm{B}~\|~\bm{B}_0) \\
    &\quad + \lambda(\alpha\|\bm{X}_1\|_1 + (1-\alpha)R(\bm{X}_2)) \\
    &\quad + \rho\sideset{}{_{i=1,2}}\sum\Bigl(\langle \bm{Z}_i, \bm{B} - \bm{X}_i \rangle + B_{\phi}(\bm{B}, \bm{X}_i)\Bigr).
\end{aligned}
\end{eqnarray}
where $\bm{Z}_1$ and $\bm{Z}_2$ are dual variables, $B_{\phi}$ denotes the Bregman divergence term determined by a convex function $\phi$. 
When $\phi=\|\cdot\|_2$, $B_{\phi}(\bm{B},\bm{X}_i)=\frac{1}{2}\|\bm{B}-\bm{X}_i\|_F^2$.
When $\phi$ is the entropy function, $B_{\phi}(\bm{B},\bm{X}_i)=\text{KL}(\bm{B}\|\bm{X}_i)$. 

\begin{figure}[t]
    \centering   
    \includegraphics[width=0.9\linewidth]{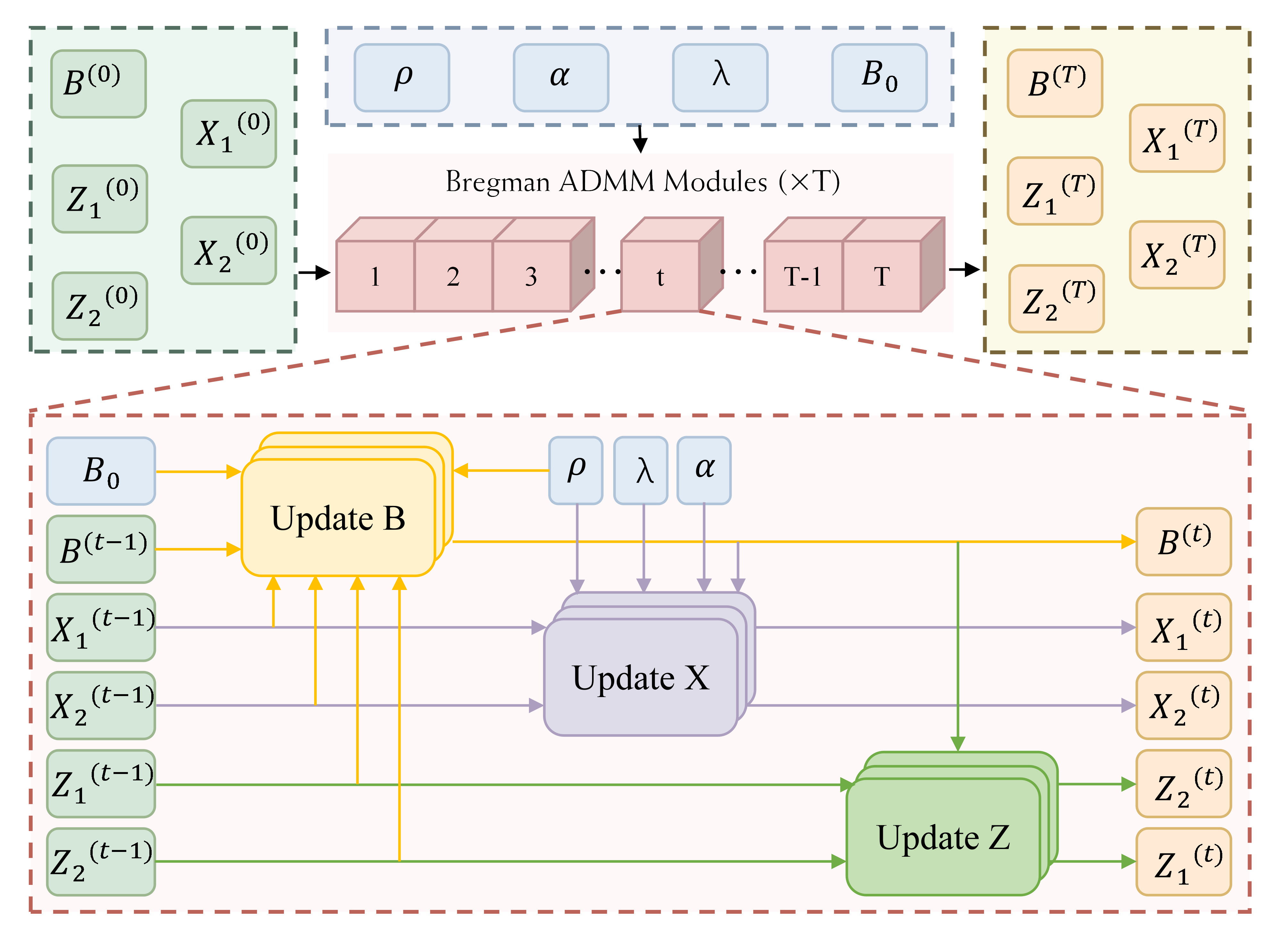}
    \caption{An illustration of the feed-forward step of our Bregman ADMM module.}
    \label{fig:badmm}
\end{figure}

After initializing $\bm{R}^{(0)}=\bm{X}_1^{(0)}=\bm{X}_2^{(0)}=\bm{B}_0$ and $\bm{Z}_1^{(0)}=\bm{Z}_2^{(0)}=\bm{0}_{N\times N}$, we update the variables iteratively by alternating optimization.
In the $t$-th iteration, we have
\begin{itemize}
    \item \textbf{Update $\bm{B}$} by solving the following problem:
    \begin{eqnarray*}
    \begin{aligned}
        \sideset{}{}\min_{\bm{B}\in\Omega} \text{KL}(\bm{B}\|\bm{B}_0)+\rho\sideset{}{}\sum_{i=1,2} \Big( \langle \bm{Z}_i^{(t)}, \bm{B} \rangle + \text{KL}(\bm{B}\|\bm{X}_i^{(t)}) \Big),
    \end{aligned}
    \end{eqnarray*}
    where the Bregman divergence is set to be KL-divergence.
    This problem has a closed form solution:
    \begin{eqnarray*}
        \bm{B}^{(t+1)} = \sigma_r \Big( \frac{\log \bm{B}_0+\rho \sum_{i=1,2} (\log \bm{X}_i^{(t)} - \bm{Z}_i^{(t)})}{1+2\rho}\Big).
    \end{eqnarray*}
    \item \textbf{Update $\bm{X}_1$ and $\bm{X}_2$} via soft-thresholding: Setting the Bregman divergence to be the squared $\ell_2$-norm, we have
    \begin{eqnarray*}
    \begin{aligned}
        &\sideset{}{}\min_{\bm{X}_1} \frac{\rho}{2} \|\bm{X}_1 - \bm{B}^{(t+1)}\|_2^2 - \rho \langle \bm{Z}_1^{(t)}, \bm{X}_1 \rangle  + \lambda \alpha \|\bm{X}_1\|_1\\
        &\Rightarrow~\bm{X}_1^{(t+1)} =  S_{\frac{\lambda \alpha}{\rho}} (\bm{B}^{(t+1)} + \bm{Z}_1^{(t)}), 
    \end{aligned}
    \end{eqnarray*}
    where $S_{\tau}(a)=\text{sign}(a)(|a|-\tau)_+$ is the elementwise soft-thresholding function. 
    \begin{eqnarray*}
    \begin{aligned}
        &\sideset{}{}\min_{\bm{X}_2} \frac{\rho}{2} \|\bm{X}_2 - \bm{B}^{(t+1)}\|_2^2 - \rho \langle \bm{Z}_2^{(t)}, \bm{X}_2 \rangle + \lambda \alpha R(\bm{X}_2)\\
        &\Rightarrow~
        \begin{cases}
        \bm{X}_2^{(t+1)}= 
        \bm{U} S_{\frac{\lambda (1-\alpha)}{\rho}} (\bm{\Sigma}) \bm{V}^{\top},&R(\cdot)=\|\cdot\|_*,\\
        \bm{x}_{n,2}^{(t+1)}=\tau_n S_{\frac{\lambda \alpha}{\rho}} (\bm{b}_n^{(t+1)} + \bm{z}_{n,2}^{(t)}),&R(\cdot)=\|\cdot\|_{1,2},
        \end{cases}
    \end{aligned}
    \end{eqnarray*}
    where we denote $\bm{x}_{n,2}^{(t+1)}$, $\bm{z}_{n,2}^{(t)}$, and $\bm{b}_n^{(t+1)}$ as the $n$-th column of $\bm{X}_2^{(t+1)}$, $\bm{Z}_{2}^{(t)}$, and $\bm{B}^{(t+1)}$, respectively. 
    $\tau_n=(1-\frac{(1-\alpha)\lambda}{\rho\|S_{\frac{\lambda \alpha}{\rho}} (\bm{b}_{n}^{(t+1)} + \bm{z}_{n,2}^{(t)})\|_2})_+$. 
    $\bm{U}\bm{\Sigma}\bm{V}^{\top}$ is the singular value decomposition (SVD) of $\bm{B}^{(t+1)}+\bm{Z}_2^{(t)}$. 
    The closed-form solutions are derived based on the implementations of $R(\cdot)$.
    \item \textbf{Update $\bm{Z}_1$ and $\bm{Z}_2$} by the following step:
    \begin{eqnarray}
    \begin{aligned}
    \bm{Z}_1^{(t+1)} &= \bm{Z}_1^{(t)} + (\bm{B}^{(t+1)} - \bm{X}_1^{(t+1)}), \\
    \bm{Z}_2^{(t+1)} &= \bm{Z}_2^{(t)} + (\bm{B}^{(t+1)} - \bm{X}_2^{(t+1)}).
    \end{aligned}
    \end{eqnarray}
\end{itemize}
Repeating the above updates $T$ times till the objective function converges, we obtain the final event branch matrix. 

For Hawkes process, we can apply the above algorithm directly in the E-step and obtain a structured responsibility matrix accordingly.
For neural TPPs, we follow the work in~\cite{xu2023regularized}, unrolling the iterations as $T$ layers.
Accordingly, the proposed Bregman ADMM module can be implemented as the stack of the $T$ layers, as illustrated in Figure~\ref{fig:badmm}. 
Plugging the Bregman ADMM module into the attention layer leads to a new attention mechanism. 
Note that, the $T$ layers are involved in the back-propagation when learning the neural TPPs. 
When $R(\cdot)=\|\cdot\|_*$, we detach the gradient of $\bm{X}_2$ such that the SVD operation will not be considered and the memory cost can be saved.

\subsection{Comparisons with Existing Methods}
Our BADMM module is applicable to improve both traditional EM paradigms and Transformer-based TPP models for inferring event branches. 
To our knowledge, another plug-and-play module driven by optimization algorithm and with a similar functionality is the Sinkhorn-based module proposed in~\cite{mena2020sinkhorn,sander2022sinkformers}. 
Specifically, focusing on the EM algorithm of Gaussian mixture models, the work in~\cite{mena2020sinkhorn} revisits the E-step of the algorithm and computes the responsibility matrices by the Sinkhorn-scaling algorithm~\cite{sinkhorn1967concerning}. 
Similarly, the Sinkformer in~\cite{sander2022sinkformers} improves the attention layer of Transformer encoder, applying the Sinkhorn-scaling algorithm to derive attention maps. 
These methods impose a doubly-stochastic constraint on the responsibility matrices and attention maps, treating the matrices as entropic optimal transport plans~\cite{cuturi2013sinkhorn}. 

Note that although the Sinkhorn-based module can also derive structured matrices, it is inapplicable for inferring event branches of TPPs. 
For the EM algorithm of Hawkes processes, the responsibility matrix is a lower-triangular matrix.
Applying the Sinkhorn-scaling algorithm to the matrix leads to a diagonal matrix.
It means that all the events are independent, which is unreasonable in practice.
For the attention maps in neural TPPs, although we can apply the Sinkhorn-based attention layer first and then mask the doubly-stochastic attention map to a lower-triangular matrix, this operation breaks the row-wise normalization structure of the attention map.
As a result, we can no longer interpret the attention map as the posterior probabilities of historical events given the current ones. 

\section{Experiments}
We conducted comprehensive experiments on both synthetic and real-world datasets and evaluate the usefulness of our module. 
In addition, we analyzed the interpretabilty of inferred event branches and demonstrated the robustness of our module to its hyperparameters. 
All the experiments are run on two 3090 GPUs, and we record each method's averaged performance and standard deviation in three trials.

\subsection{Implementation Details}
\subsubsection{Datasets}
We apply one synthetic event sequence dataset and four real-world ones in our experiments, whose statistics are shown in Table~\ref{tab:data}.
The synthetic dataset \textbf{Conttime}~\cite{mei2017neural} is simulated by Hawkes process, while the real-world datasets, including \textbf{Retweet}~\cite{zhao2015seismic}, \textbf{StackOverflow (SO)}~\cite{jure2014snap}, \textbf{Amazon}~\cite{xue2023easytpp}, and  \textbf{Taobao}~\cite{xue2023easytpp}, contain user behaviors on different platforms, which are commonly used for evaluating TPP models.
In addition, to evaluate the rationality of inferred event branches, we consider the dialogue data in the movie ``\textbf{12 Angry Men}''~\cite{zhang2018started}, in which each juror is treated as an event type and the timestamped sentences of the jurors are events.

\begin{table}[t]
\caption{The statistics of datasets}
\label{tab:data}
    \centering
    \small{
    \begin{tabular}{c|c|c|c}
    \hline\hline 
    Dataset & \# Event Types & \# Sequences & Max. length \\
    \hline 
    Conttime & 5 & 10,000 & 100 \\
    Taobao &17 & 2,000 & 94 \\
    Retweet & 3 & 24,000 & 264 \\
    StackOverflow & 22  & 6,633 & 736  \\
    Amazon & 16 & 12,556 & 282\\
    12 Angry Men & 12 & 1 & 587\\
    \hline\hline
    \end{tabular}
    }
\end{table}

\begin{table*}[t]
  \centering
  \begin{small}
  \caption{Comparisons for various methods on model performance.}
  \small{
  \tabcolsep=2.5pt
    \begin{tabular}{c|c|cc|cc|cc|cc|cc}
    \toprule \toprule
    \multirow{2}{*}{Model} & 
    \multirow{2}{*}{Method} &
    \multicolumn{2}{c|}{Taobao} & 
    \multicolumn{2}{c|}{Retweet} & 
    \multicolumn{2}{c|}{StackOverflow} &
    \multicolumn{2}{c|}{Amazon} &
    \multicolumn{2}{c}{Conttime}\\
    & & 
    ELL $\uparrow$ & ACC $\uparrow$ &  
    ELL $\uparrow$ & ACC $\uparrow$ &  
    ELL $\uparrow$ & ACC $\uparrow$ & 
    ELL $\uparrow$ & ACC $\uparrow$ & 
    ELL $\uparrow$ & ACC $\uparrow$ \\
    \midrule

    \multirow{3}{*}{HP} & 
    \multirow{1}{*}{Classic EM} & 
    -0.262$_{\text{0.014}}$ & 0.463$_{\text{0.001}}$ & 
    -9.022$_{\text{0.017}}$ & 0.459$_{\text{0.008}}$ & 
    -3.042$_{\text{0.089}}$ & 0.439$_{\text{0.002}}$ & 
    -2.561$_{\text{0.000}}$ & 0.355$_{\text{0.002}}$ & 
    -1.076$_{\text{0.001}}$ & 0.340$_{\text{0.001}}$ \\ 
    &\multirow{1}{*}{BADMM$_{1,2}$} & 
    \hspace{0.1cm}0.220$_{\text{0.000}}$ & 0.476$_{\text{0.001}}$ & 
    -8.902$_{\text{0.001}}$ & 0.564$_{\text{0.000}}$ & 
    -2.758$_{\text{0.012}}$ & 0.452$_{\text{0.002}}$ & 
    -2.561$_{\text{0.000}}$ & 0.366$_{\text{0.001}}$ & 
    -1.073$_{\text{0.000}}$ & 0.340$_{\text{0.001}}$ \\ 
   &\multirow{1}{*}{BADMM$_*$} &
   
    \textbf{\hspace{0.1cm}0.223}$_{\text{0.000}}$ & \textbf{0.480}$_{\text{0.001}}$ & 
    \textbf{-8.902}$_{\text{0.000}}$ & \textbf{0.565}$_{\text{0.000}}$ & 
    \textbf{-2.744}$_{\text{0.010}}$ & \textbf{0.454}$_{\text{0.002}}$ & 
    \textbf{-2.561}$_{\text{0.000}}$ & \textbf{0.368}$_{\text{0.000}}$ & 
    \textbf{-1.073}$_{\text{0.000}}$ & \textbf{0.340}$_{\text{0.001}}$ \\ 
    \midrule
    
    \multirow{4}{*}{SAHP} & 
    \multirow{1}{*}{Softmax} & 
    -9.879$_{\text{0.856}}$ & 0.434$_{\text{0.000}}$ & 
    -7.878$_{\text{0.045}}$ & 0.537$_{\text{0.001}}$ & 
    -3.614$_{\text{0.247}}$ & 0.353$_{\text{0.012}}$ & 
    -3.223$_{\text{0.031}}$ & 0.295$_{\text{0.052}}$ & 
    -1.416$_{\text{0.104}}$ & 0.264$_{\text{0.037}}$ \\ 
    & \multirow{1}{*}{Sinkhorn} & 
    -9.767$_{\text{0.552}}$ & 0.434$_{\text{0.000}}$ & 
    -7.702$_{\text{0.488}}$ & 0.541$_{\text{0.005}}$ & 
    -3.448$_{\text{0.097}}$ & 0.356$_{\text{0.010}}$ & 
    -3.137$_{\text{0.283}}$ & 0.306$_{\text{0.029}}$ & 
    -1.363$_{\text{0.181}}$ & 0.272$_{\text{0.026}}$\\ 
    &\multirow{1}{*}{BADMM$_{1,2}$} & 
    -9.616$_{\text{0.485}}$  & 0.434$_{\text{0.000}}$ & 
    -7.802$_{\text{0.462}}$ & 0.538$_{\text{0.001}}$ & 
    -3.625$_{\text{0.448}}$ & 0.356$_{\text{0.012}}$ & 
    -3.133$_{\text{0.271}}$ & 0.299$_{\text{0.036}}$ & 
    -1.305$_{\text{0.102}}$ & \textbf{0.283}$_{\text{0.029}}$ \\ 
    & \multirow{1}{*}{BADMM$_*$} & 
    \textbf{-9.594}$_{\text{0.119}}$ & \textbf{0.434}$_{\text{0.000}}$ & 
    \textbf{-7.686}$_{\text{0.090}}$ & \textbf{0.542}$_{\text{0.003}}$ & 
    \textbf{-3.374}$_{\text{0.062}}$ & \textbf{0.365}$_{\text{0.003}}$ & 
    \textbf{-3.127}$_{\text{0.282}}$ & \textbf{0.307}$_{\text{0.009}}$ & 
    \textbf{-1.235}$_{\text{0.086}}$ & 0.278$_{\text{0.022}}$ \\ 
    \midrule

    \multirow{4}{*}{THP} & 
    \multirow{1}{*}{Softmax} & 
    0.170$_{\text{0.006}}$ & 0.436$_{\text{0.000}}$ & 
    -9.373$_{\text{0.716}}$ & 0.511$_{\text{0.013}}$ & 
    -0.698$_{\text{0.007}}$ & 0.450$_{\text{0.002}}$ & 
    0.542$_{\text{0.001}}$& 
    0.351$_{\text{0.001}}$& 
    -0.390$_{\text{0.043}}$& 0.696$_{\text{0.051}}$ \\ 
     & \multirow{1}{*}{Sinkhorn} & 
    0.178$_{\text{0.007}}$& 0.436$_{\text{0.000}}$ & 
    -11.020$_{\text{0.134}}$ & 0.525$_{\text{0.008}}$ & 
    -0.700$_{\text{0.010}}$ & 0.433$_{\text{0.004}}$ & 
    0.288$_{\text{0.025}}$ & 
    0.331$_{\text{0.022}}$ & 
    -0.394$_{\text{0.039}}$ & 0.633$_{\text{0.036}}$ \\ 
    &\multirow{1}{*}{BADMM$_{1,2}$} & 
    0.190$_{\text{0.013}}$& 0.436$_{\text{0.000}}$& 
     -8.961$_{\text{0.233}}$ & 0.535$_{\text{0.003}}$ & 
    -0.696$_{\text{0.008}}$& 0.453$_{\text{0.003}}$& 
    0.542$_{\text{0.001}}$&
    0.355$_{\text{0.001}}$& 
    -0.390$_{\text{0.035}}$& 0.727$_{\text{0.031}}$\\ 
    & \multirow{1}{*}{BADMM$_*$} &
    \textbf{0.193}$_{\text{0.012}}$& \textbf{0.436}$_{\text{0.000}}$ & 
     \textbf{-8.753}$_{\text{0.046}}$ & \textbf{0.538}$_{\text{0.001}}$ & 
    \textbf{-0.696}$_{\text{0.002}}$& \textbf{0.453}$_{\text{0.002}}$& 
   \textbf{0.543}$_{\text{0.000}}$ & \textbf{0.356}$_{\text{0.001}}$ &  \textbf{-0.385}$_{\text{0.041}}$& \textbf{0.748}$_{\text{0.025}}$\\ 
    \bottomrule \bottomrule
    \end{tabular}
  }
  \label{tab:cmp}%
\end{small}
\end{table*}%

\subsubsection{Baselines} 
We plug our BADMM module into existing TPPs and evaluate its impacts accordingly. 
Here, we consider the classic Hawkes process (\textbf{HP} in~\cite{zhou2013learning}) and two Transformer-based TPPs (\textbf{THP} in~\cite{zuo2020transformer} and \textbf{SAHP} in~\cite{zhang2020self}). 
For HP, we apply our BADMM module to estimate its responsibility matrix in the EM algorithm.
For THP and SAHP, we replace their self-attention layer with our BADMM module. 
The baselines of our method include: $i)$ the HP learned by EM algorithm, $ii)$ the HP-based source tracking (\textbf{HPST}) in~\cite{zhang2018started}, $iii)$ the THP and SAHP learned by SGD, and $iv)$ the THP and SAHP applying the Sinkhorn-based attention layers in~\cite{sander2022sinkformers}. 

\subsubsection{Hyperparameter settings} 
For our BADMM module, we consider two implementations, denoted as $\textbf{BADMM}_*$ (i.e., $R(\cdot)=\|\cdot\|_*$) and $\textbf{BADMM}_{1,2}$ (i.e., $R(\cdot)=\|\cdot\|_{1,2}$), respectively.
The weight $\rho$ of Lagrangian term only affects the convergence rate, so we simply set it to be $1$ in our experiments.
For the weight of regularizer, we set $\alpha\in (0, 1)$ and $\lambda\in\{0.01, 0.1, 1, 10, 100\}$. 
Applying grid search, we find their optimal configurations.
For the number of iterations $T$, we follow the setting of Sinkhorn-based module in~\cite{sander2022sinkformers} and set $T=2$, which achieves a trade-off between the model complexity and capability.

\subsubsection{Evaluation metrics}

When evaluating each learned model in a testing set, we record $i$) the log-likelihood per event (\textbf{ELL}) and $ii$) the prediction accuracy of event types (\textbf{ACC}). 
In addition, to evaluate the rationality of inferred event branches, we visualize the transition matrices learned by different methods and check the learned important events and their triggering patterns qualitatively.

\subsection{Effectiveness and Rationality}
\subsubsection{Impacts on Model Performance}
We conducted a thorough evaluation of BADMM-enhanced methods against various baselines across multiple datasets. 
The results in Table~\ref{tab:cmp} demonstrate the consistent effectiveness of our BADMM module across all datasets and backbone models.
In most situations, applying our BADMM module significantly improves the performance of both classical and neural TPPs.
In our opinion, this phenomenon is because the event sequences in practice are driven by hidden branching processes, so that inferring event branches during training does not do harm to the learning of TPPs.
Moreover, the existing TPPs, especially those neural ones, may suffer from the over-fitting issue during training.
Our BADMM module imposes structural constraints on their transition matrices, regularizing model parameters implicitly to mitigate the issue.

\begin{figure}[t]
    \centering
    \subfigure[$\bm{R}$ in HP]{
    \tabcolsep=0pt
    \begin{tabular}{ccc}
    \includegraphics[height=2.1cm]{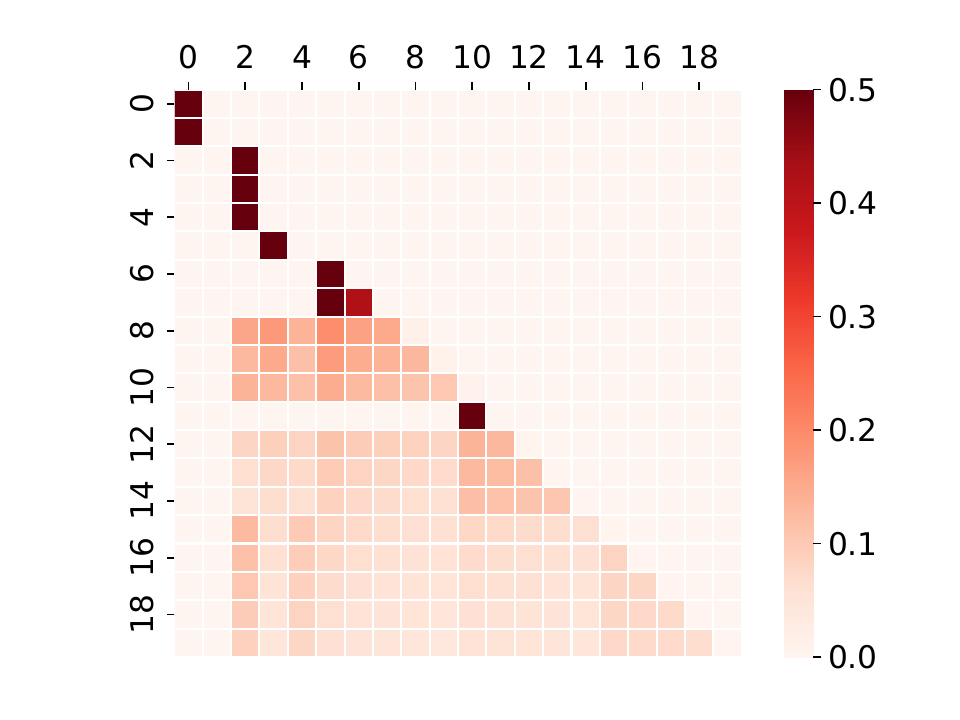} &
    \includegraphics[height=2.1cm]{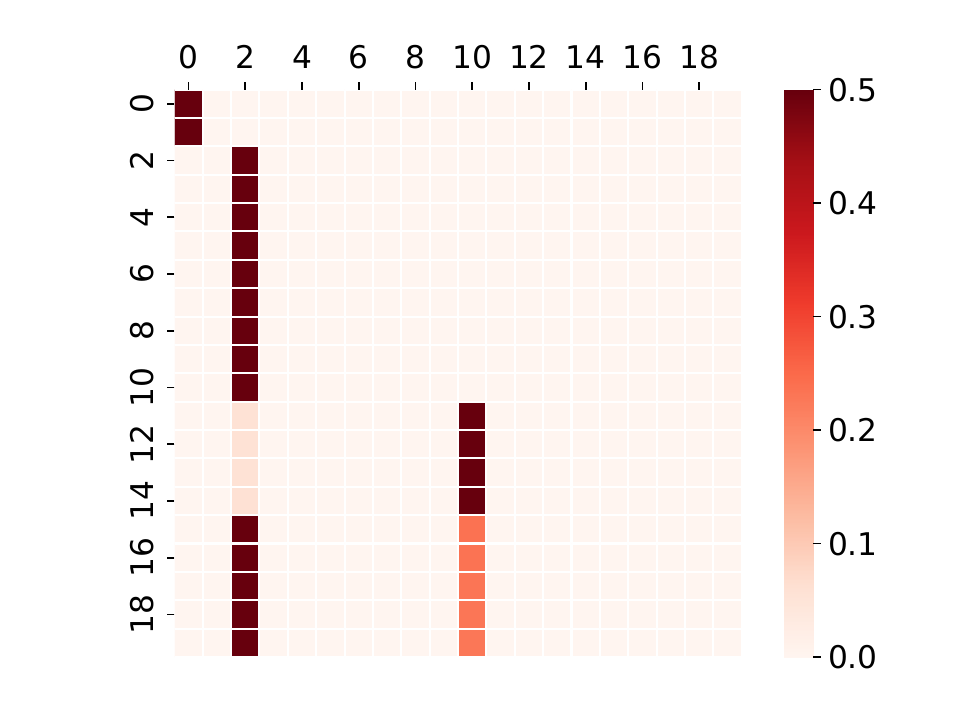} &
    \includegraphics[height=2.1cm]{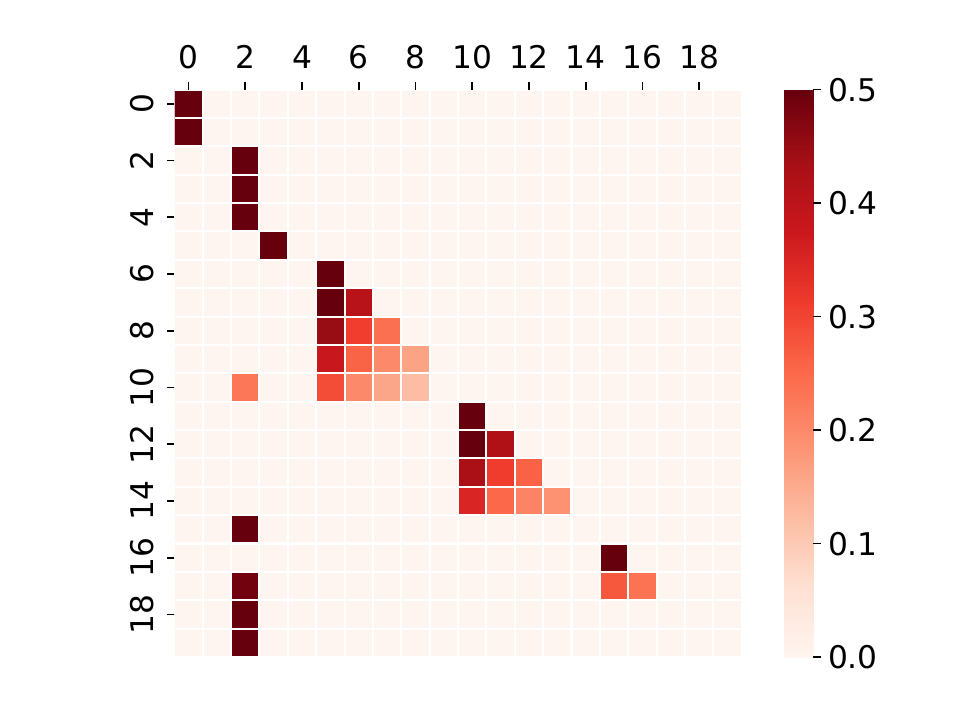} \\
    Classic EM & BADMM$_{1,2}$ & BADMM$_{*}$
    \end{tabular}
    }
    \subfigure[$\bm{\tilde{A}}$ in SAHP]{
    \tabcolsep=0pt
    \begin{tabular}{cccc}
    \includegraphics[height=1.6cm]{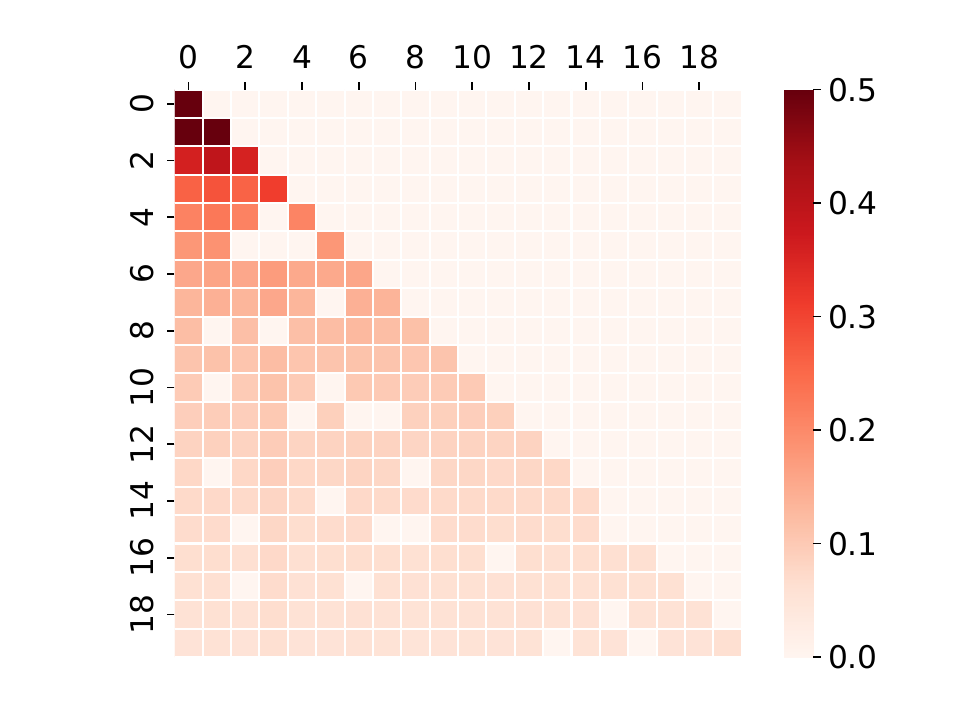} &
    \includegraphics[height=1.6cm]{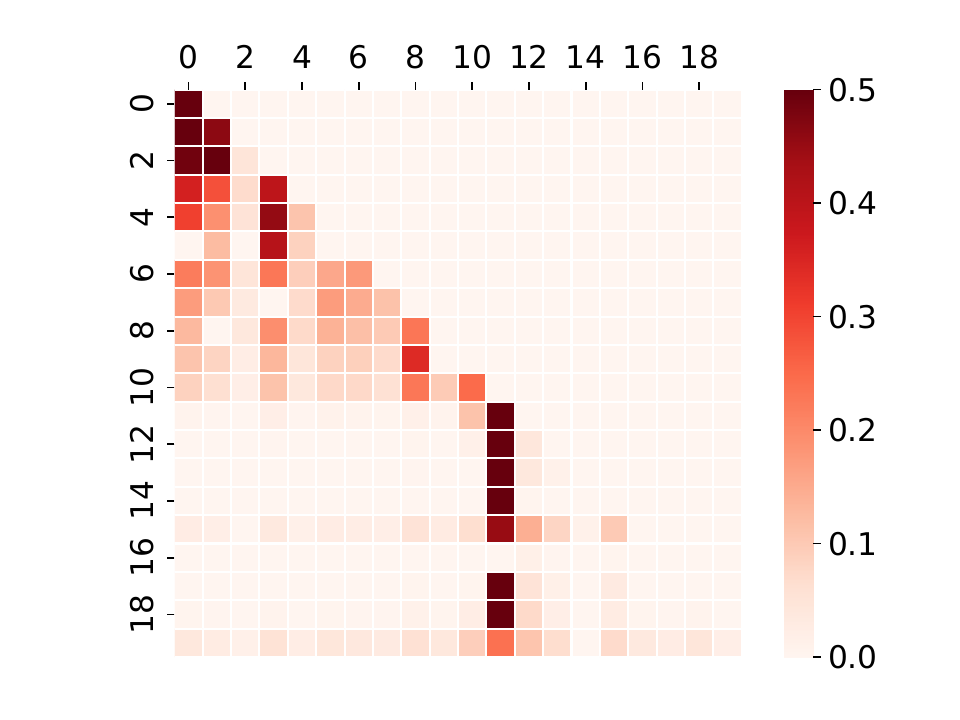} &
    \includegraphics[height=1.6cm]{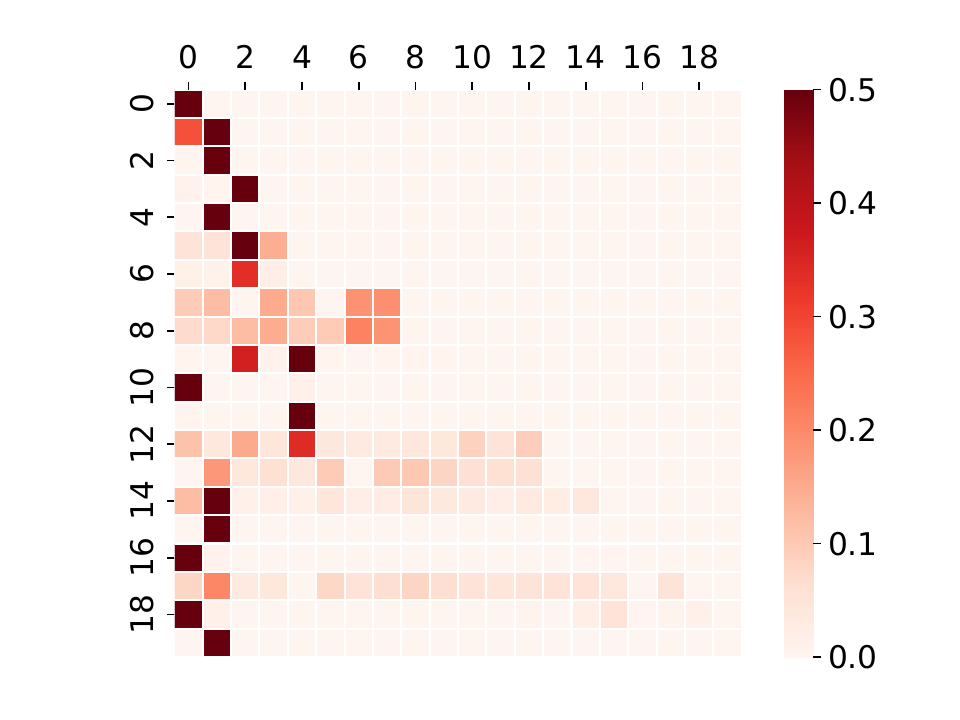} &
    \includegraphics[height=1.6cm]{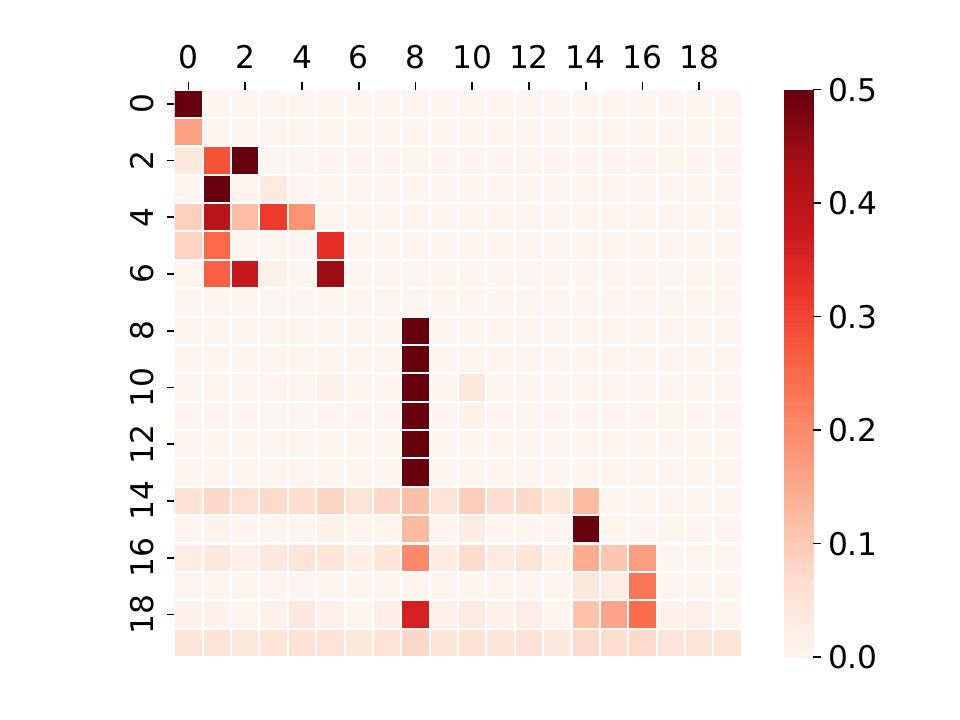}\\
    Attention & Sinkhorn & BADMM$_{1,2}$ & BADMM$_{*}$
    \end{tabular}
    }
    \subfigure[$\bm{\tilde{A}}$ in THP]{
    \tabcolsep=0pt
    \begin{tabular}{cccc}
    \includegraphics[height=1.6cm]{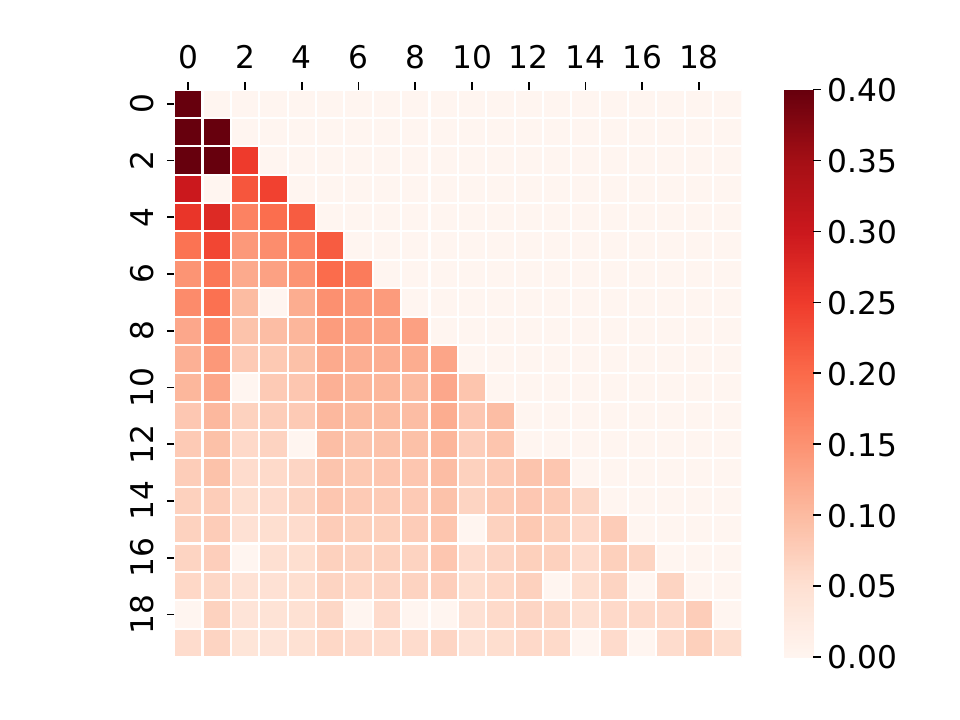} &
    \includegraphics[height=1.6cm]{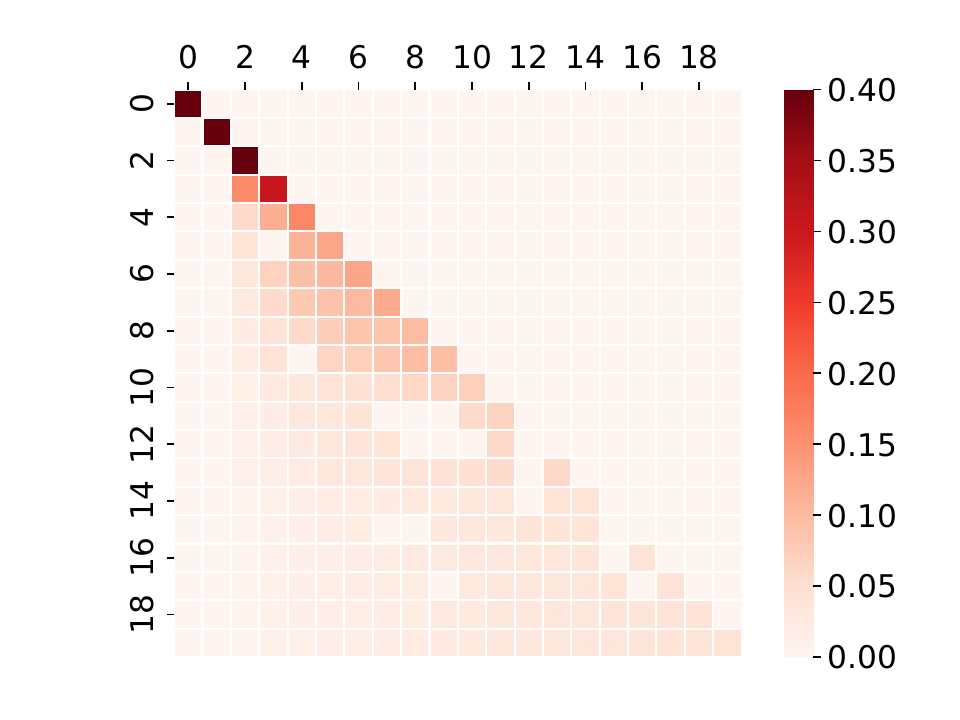} &
    \includegraphics[height=1.6cm]{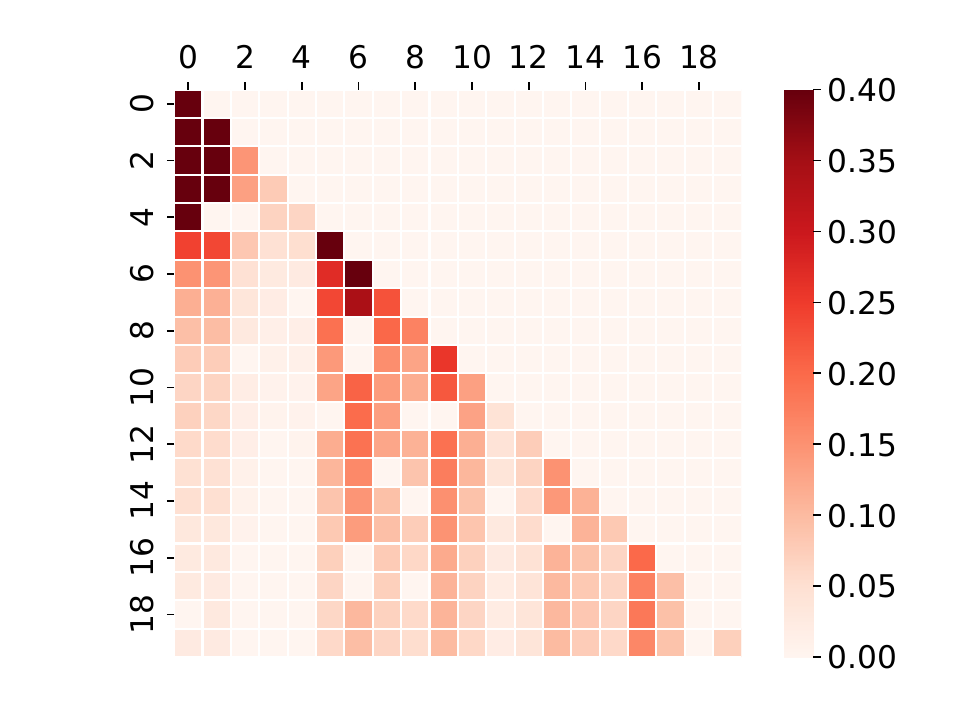} &
    \includegraphics[height=1.6cm]{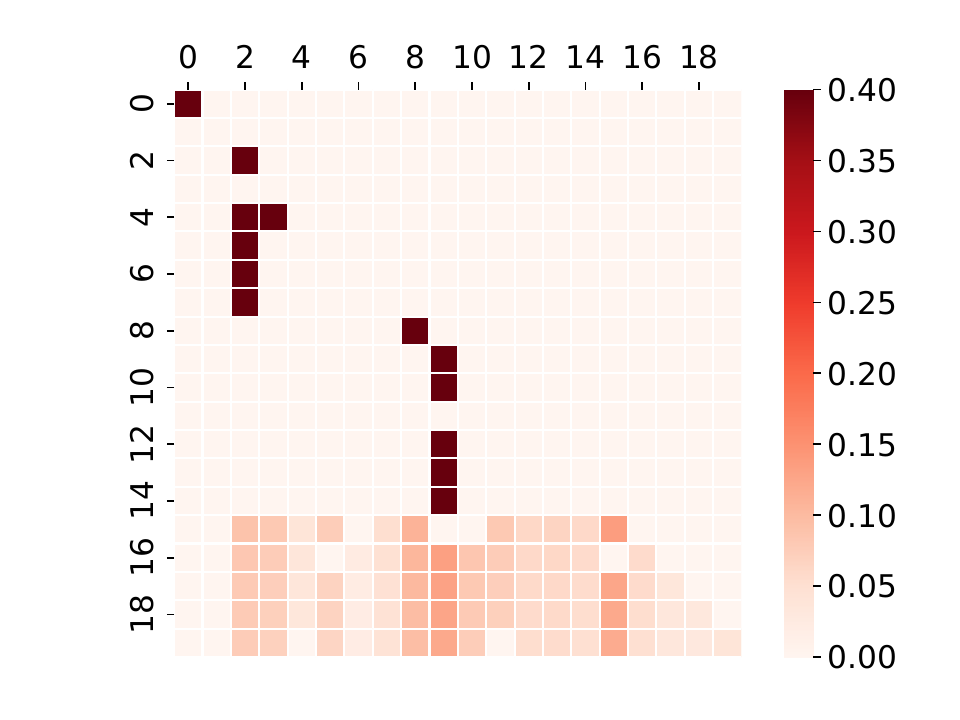}\\
    Attention & Sinkhorn & BADMM$_{1,2}$ & BADMM$_{*}$
    \end{tabular}
    }
    \caption{Visualization of inferred transition matrices. The event sequence is from the Conttime dataset.}
    \label{fig:visualize}   
\end{figure}  

\begin{table}[t]
\caption{Top-5 most influential jurors in ``12 Angry Men''}
\label{tab:testing_result}
    \centering
    \small{
    \tabcolsep=4pt
    \begin{tabular}{c|c|c|ccc}
    \hline\hline
    \multirow{2}{*}{\textbf{Rank}} 
    & \multirow{2}{*}{\textbf{GPT-4o}} 
    & \multirow{2}{*}{\textbf{HPST}}
    & \multicolumn{3}{c}{\textbf{BADMM}}\\
    &
    &
    & HP
    & THP 
    & SAHP \\
    \hline
    
    \multirow{1}{*}{1}
    & Juror 8
    & Juror 8
    & Juror 8 
    & Juror 8
    & Juror 8  \\

    \multirow{1}{*}{2}
    & Juror 3
    & Juror 3
    & Juror 3 
    & Juror 3 
    & Juror 3  \\

    \multirow{1}{*}{3}
    & Juror 7
    & Juror 7
    & Juror 7
    & Juror 7
    & Juror 7 \\

    \multirow{1}{*}{4}
    & Juror 10
    & Juror 1
    & Juror 10
    & Juror 10
    & Juror 4 \\

    \multirow{1}{*}{5}
    & Juror 4
    & Juror 10
    & Juror 12
    & Juror 1
    & Juror 10 \\
    \hline
    \hline
    \end{tabular}
    }\label{tab:movie}
\end{table}

\begin{figure}[t]
    \centering
    \subfigure[$\bm{\tilde{A}}$ derived by Softmax]{
    \includegraphics[height=3.1cm]{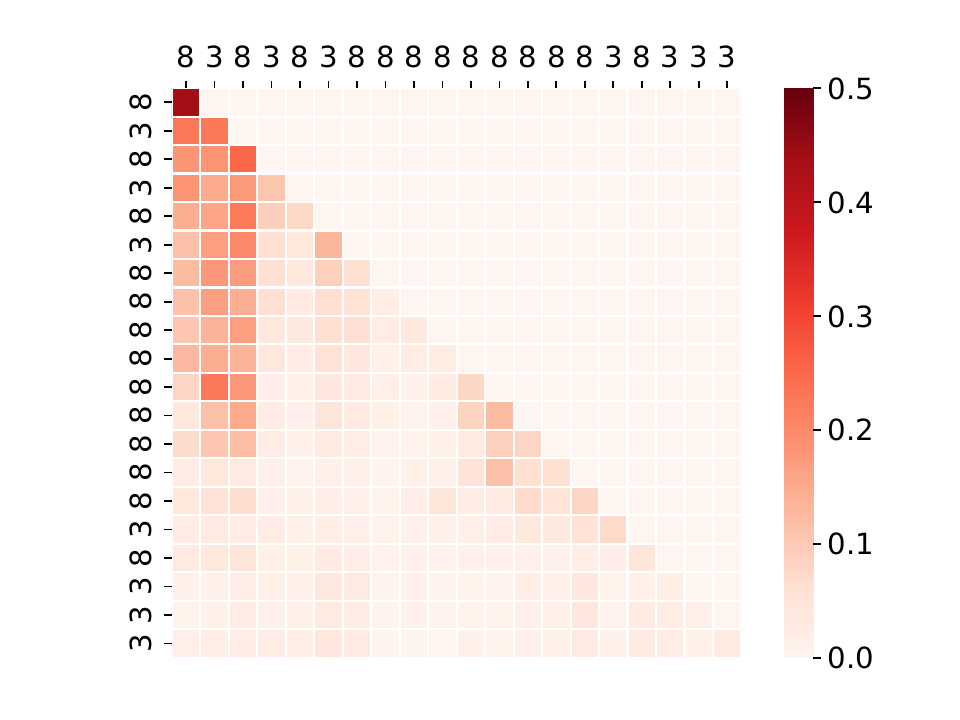}\label{fig:thpsoftmax}
    }
    \subfigure[$\bm{\tilde{A}}$ derived by BADMM$_{*}$]{
    \includegraphics[height=3.1cm]{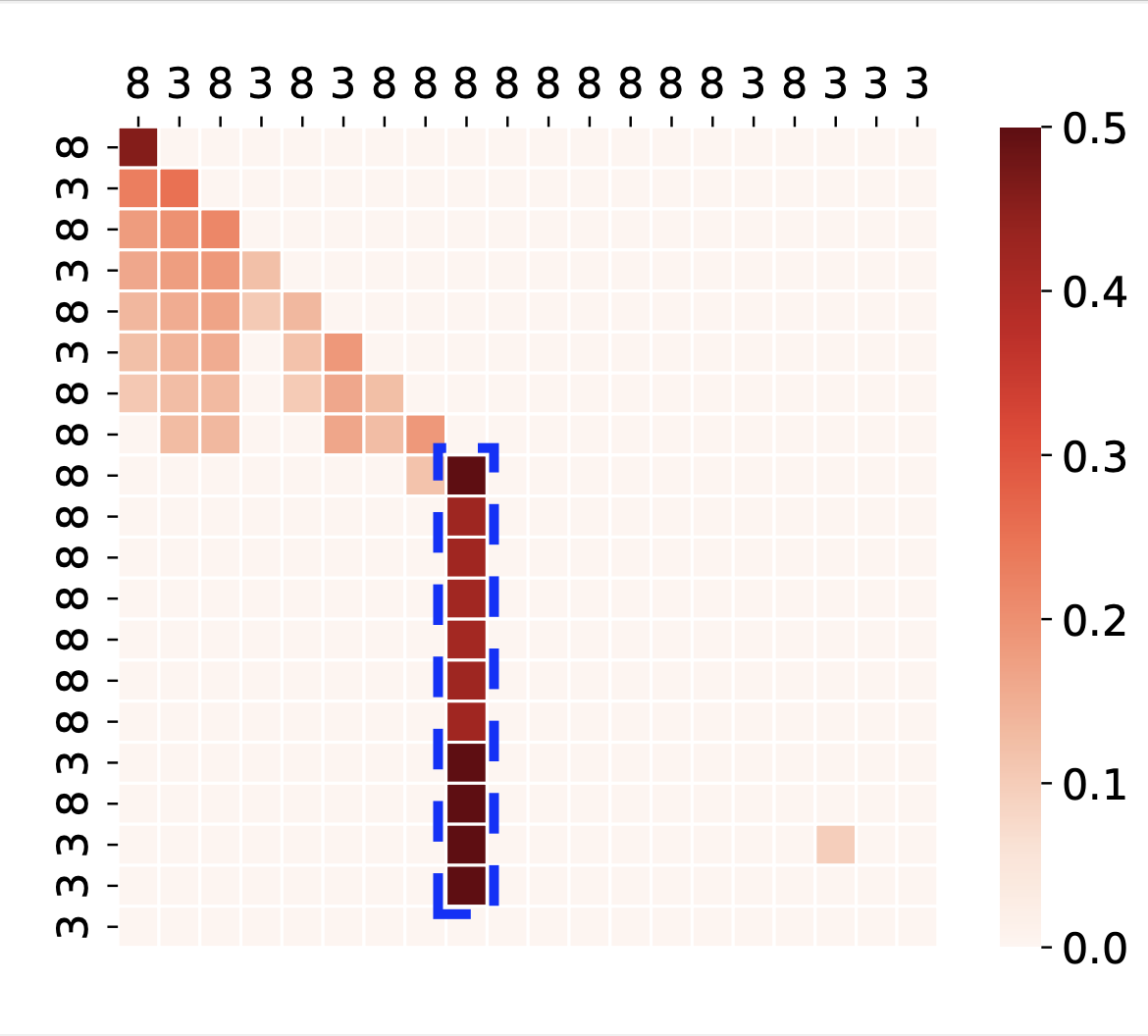}\label{fig:sahpbadmm}
    }
    \subfigure[Triggering patterns among key dialogues]{
    \includegraphics[height=5.2cm]{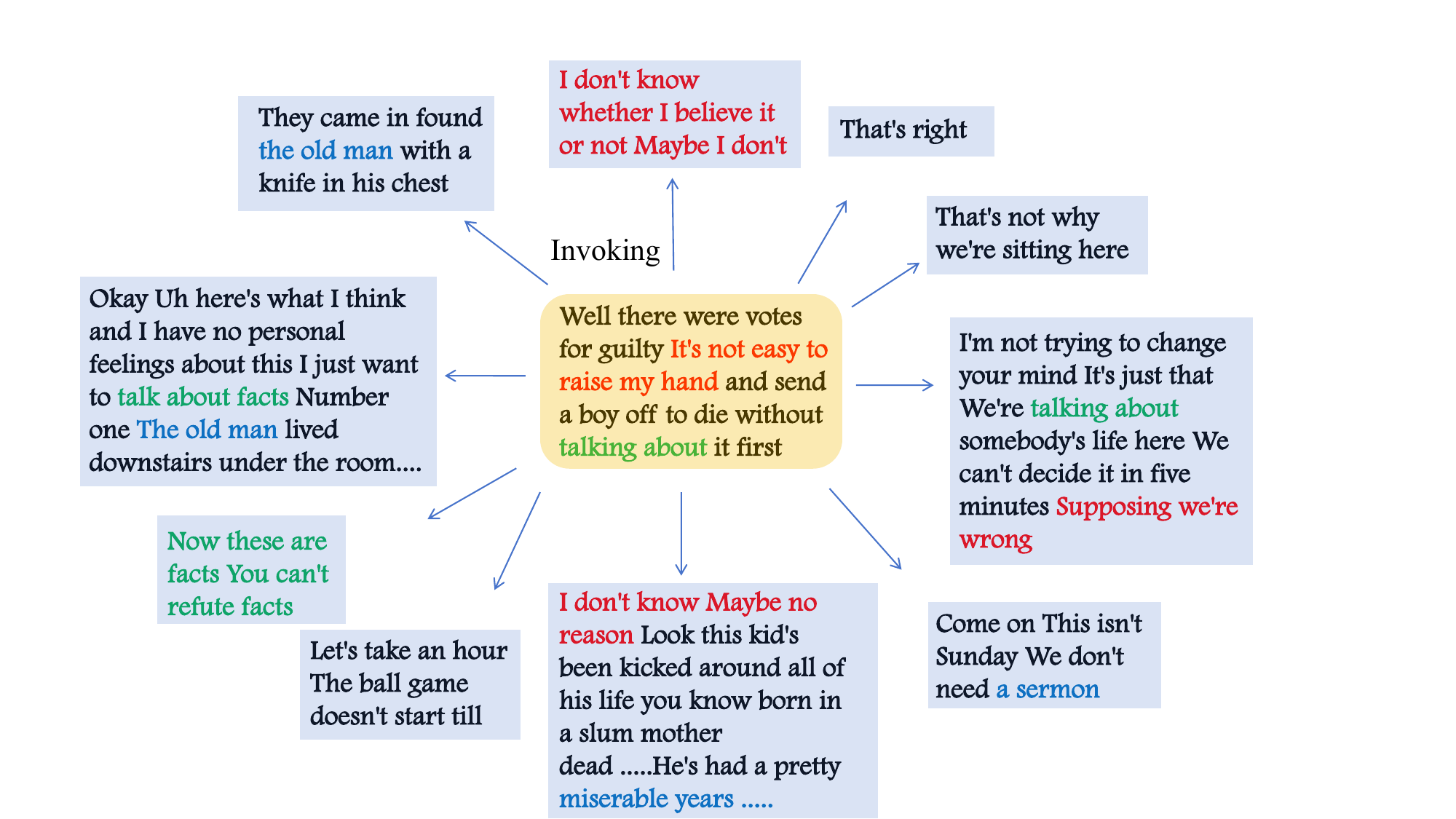}\label{fig:12a}
    }
    \caption{(a, b) Inferred event branches for ``12 Angry Men''. 
    (c) The branch triggered by the key sentence (corresponding to the blue region in (b)).}
    \label{fig:movie}
\end{figure}

\begin{figure}[t]
    \centering
    \subfigure[ACC of HP]{
    \includegraphics[height=2.0cm]{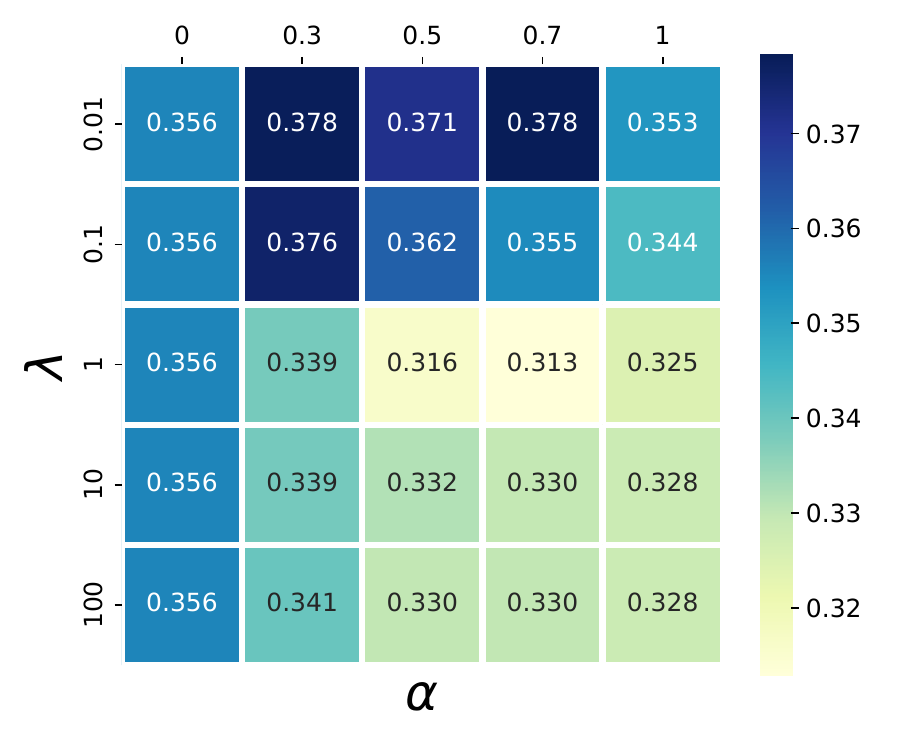}
    }
    \subfigure[ACC of SAHP]{
    \includegraphics[height=2.0cm]{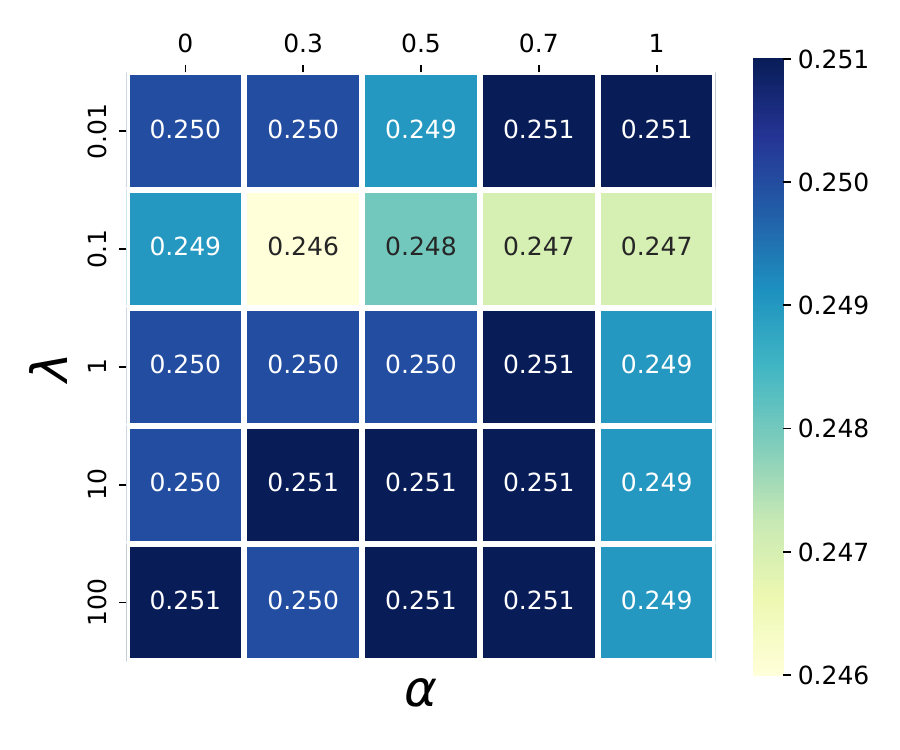}
    }
    \subfigure[Iterations v.s. ACC]{
    \includegraphics[height=2.0cm]{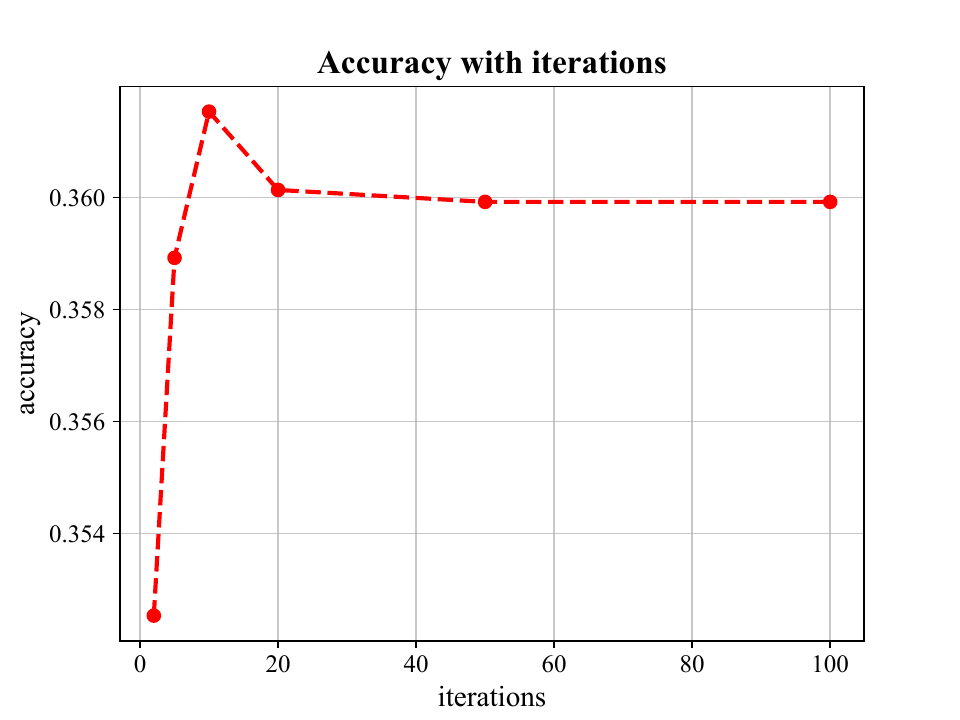}
    }
    \caption{Robustness test of our BADMM module.}
    \label{fig:robust}   
\end{figure}  

\subsubsection{Qualitative comparisons}
In Figure~\ref{fig:visualize}, we visualize the transition matrices inferred by different methods. 
We can find that the original transition matrices are over-smoothed, which contain too many nonzero elements. 
In contrast, applying our BADMM module indeed leads to sparse and low-rank transition matrices for both classic and neural TPPs, which enhances the structures of transition matrices significantly.
For neural TPPs, we further compare the transition matrices derived by our BADMM module with those derived by the Sinkhorn module~\cite{sander2022sinkformers}. 
The Sinkhorn module first derives a doubly stochastic matrix and then masks it to a lower triangular matrix.
Although it can achieve low-rank and sparse transition matrices to some extent, it breaks the row-normalization constraint of the transition matrix and thus results in inferior performance compared to our method (as shown in Table~\ref{tab:cmp}). 

In addition, the two implementations of BADMM module often lead to different event branches. 
Because of using sparse group-lasso regularization, BADMM$_{1,2}$ tends to very sparse transition matrices in which only some columns have dense nonzero elements. 
It means that BADMM$_{1,2}$ attributes the generation of the event sequence from few globally-significant events.
On the other side, BADMM$_{*}$ preserves the structures of the initial transition matrices better and tends to identify the events having local impacts on its subsequent events. 
According to the results in Table~\ref{tab:cmp}, we can find that BADMM$_{*}$ often outperforms BADMM$_{1,2}$ on data fitness and prediction accuracy, which means that the transition matrices of BADMM$_{*}$ are often associated with the models with low risks of misspecification. 
Considering this fact, we think BADMM$_{*}$ can generate more reasonable event branches and reveal triggering patterns among events better.
In the following experiments, we mainly test the performance of BADMM$_*$.

\subsubsection{Rationality of Inferred Event Branches}
For the dialogue in the movie ``12 Angry Men'', we follow the HPST method~\cite{zhang2018started}, learning a transformer hawkes process and inferring a transition matrix for the jurors' sentences. 
According to the learned transformer hawkes process, we can rank the 12 jurors in the movie based on the overall impacts of their sentences (i.e., $\bm{1}^{\top}\bm{B}\bm{S}$, where $\bm{1}^{\top}\bm{B}\in\mathbb{R}^N$ records the impacts of $N$ sentences and $\bm{S}\in\{0,1\}^{N\times 12}$ maps the sentences to corresponding jurors). 
Ideally, we hope that the ranking result can reflect the significance of the jurors in the movie.
Taking the ranking result derived by GPT-4o as the ground truth (which is checked by five volunteers watched the movie before), we compare the ranking results derived by the TPPs learned with our BADMM module with that achieved by HPST. 
Table~\ref{tab:movie} lists the top-5 most influential jurors proposed by different methods.
Both our method and HPST can detect the top-2 most influential jurors (i.e., Jurors 8 and 3, who insist on acquittal and conviction, respectively). 
Compared to HPST, the ranking results of our method are more consistent with that of GPT-4o. 
For example, the SAHP learned with our BADMM module considers Juror 4 in the top-5 list.
This juror appears in the list of GPT-4o but is missed by HPST. 
Note that both GPT-4o and HPST leverage textual information when ranking the jurors, while our method purely relies on the timestamps and the jurors' IDs within the event sequence.
In other words, by applying our BADMM module, we can learn reliable event branches that are consistent with those estimated by the text-based models.

To further verify the rationality of inferred event branches, we sample the rows and columns of the transition matrix relevant to Jurors 8 and 3 and visualize it in Figure~\ref{fig:movie}.
Similar to the results in Figure~\ref{fig:visualize}, we can find that compared to the Softmax operation of THP, our BADMM module can derive a transition matrix with better sparsity, whose event branches are more clear. 
In addition, based on the inferred transition matrix, we show the triggering patterns hidden the dialogue between the two jurors. 
We can find that the triggering patterns are consistent with the semantics of the sentences.

\subsection{Robustness to Hyperparameters}
The impacts of key hyperparameters ($\alpha$, $\lambda$, and the number of iterations $T$) on model performance are shown in Figure~\ref{fig:robust}. 
Specifically, Figures~\ref{fig:robust}(a, b) show the ACC of the models learned with different configurations of $\alpha$ and $\lambda$. 
The results show that our BADMM module consistently performs well within a broad range of the hyperparameter settings, indicating that our method is robust to moderate changes in these hyperparameters. 
Figure~\ref{fig:robust}(c) shows the effect of the number of iterations on the performance of our method. 
We can find that our BADMM module quickly converges within a few iterations, with marginal gains observed beyond a certain threshold. 
This rapid convergence implies that, in practical scenarios, we can implement our BADMM module using limited iterations, which reduces computational costs while maintaining high performance.

\section{Conclusion}  
We introduced a novel BADMM module, which helps infer structured event branches for various TPPs. 
By integrating this module into both classic and neural TPPs, we addressed the common over-smoothness issue of event branch inference, providing an effective and robust solution to impose  sparse and low-rank structures on transition matrices.
Our method improves the interpretability of learned TPPs and enhances their model performance.

\textbf{Limitations and future work.} 
When deriving the attention map by the BADMM module, it requires to solve an optimization problem of the transition matrix, which leads to a time-consuming inference step when predicting future events. 
In the future, we will apply online BADMM algorithm to improve the efficiency of the BADMM module. 
In addition, we plan to verify the rationality of our method from the viewpoint of Statistics and try to build connections between our BADMM module and Bayesian priors.

\section{Acknowledgments} 
This work was supported by National Natural Science Foundation (62106271, 92270110), the Fundamental Research Funds for the Central Universities, and the Research Funds of Renmin University of China. 
We also acknowledge the support provided by the fund for building world-class universities (disciplines) of Renmin University of China and by the funds from Engineering Research Center of Next-Generation Intelligent Search and Recommendation, Ministry of Education, and from Intelligent Social Governance Interdisciplinary Platform, Major Innovation \& Planning Interdisciplinary Platform for the ``Double-First Class'' Initiative, Renmin University of China.

\bibliography{main}

\end{document}